\documentclass[letterpaper, 10 pt, conference]{ieeeconf}

\IEEEoverridecommandlockouts
\pdfoutput=1 

\usepackage[utf8]{inputenc}
\usepackage[T1]{fontenc}
\usepackage{url}
\usepackage{booktabs}
\usepackage{graphicx}
\usepackage{multirow}
\usepackage{subcaption} 
\usepackage{float}
\usepackage{amsfonts}
\usepackage{amsmath}
\usepackage{nicefrac}
\usepackage{microtype}
\usepackage{xcolor}
\usepackage{tabularx}
\usepackage{ragged2e}

\usepackage[bookmarks=false, breaklinks=true, letterpaper=true, colorlinks, linkcolor=black, citecolor=black, urlcolor=black]{hyperref}

\title{MTR-VP: Towards End-to-End Trajectory Planning through Context-Driven Image Encoding and Multiple Trajectory Prediction}

\author{%
  Maitrayee Keskar\thanks{Authors M. Keskar and R. Greer are with the Machine Intelligence, Interaction, and Imagination (Mi$^3$) Laboratory at the University of California, Merced. All authors are also with the Laboratory for Intelligent \& Safe Automobiles (LISA) at the University of California, San Diego.},  Mohan Trivedi, Ross Greer
}

\begin{document}

\maketitle

\begin{abstract}
We present a method for trajectory planning for autonomous driving, learning image-based context embeddings that align with motion prediction frameworks and planning-based intention input. Within our method, a ViT encoder takes raw images and past kinematic state as input and is trained to produce context embeddings, drawing inspiration from those generated by the recent MTR (Motion Transformer) encoder, effectively substituting map-based features with learned visual representations. MTR provides a strong foundation for multimodal trajectory prediction by localizing agent intent and refining motion iteratively via motion query pairs; we name our approach MTR-VP (Motion Transformer for Vision-based Planning), and instead of the learnable intention queries used in the MTR decoder, we use cross attention on the intent and the context embeddings, which reflect a combination of information encoded from the driving scene and past vehicle states. We evaluate our methods on the Waymo End-to-End Driving Dataset, which requires predicting the agent's future 5-second trajectory in bird’s-eye-view coordinates using prior camera images, agent pose history, and routing goals. We analyze our architecture using ablation studies, removing input images and multiple trajectory output. Our results suggest that transformer-based methods that are used to combine the visual features along with the kinetic features such as the past trajectory features are not effective at combining both modes to produce useful scene context embeddings, even when intention embeddings are augmented with foundation-model representations of scene context from CLIP and DINOv2, but that predicting a distribution over multiple futures instead of a single future trajectory boosts planning performance. Our code will be made publicly available after publication.

\end{abstract}

\section{Introduction}

Autonomous driving systems must be robust to a wide array of real-world situations \cite{wang2024drivedreamer}, including rare and unstructured scenarios such as construction detours, sudden pedestrian behavior, and atypical obstacles \cite{fu2024drive}. Traditional modular pipelines, which separately process perception \cite{sun2020scalability}, prediction \cite{cui2019multimodal}, and planning \cite{hu2023planning}, have shown strong performance in common cases but may fail to generalize in these "long-tail" scenarios due to error compounding, reliance on curated intermediate representations, and imitation-learning approaches trained on majority ``typical'' data \cite{li2022coda, makansi2021exposing}. Furthermore, accurately planning in highly interactive scenarios requires reasoning about the mutual influence between the ego vehicle and other agents, a challenge addressed by methods that jointly optimize the behavior of all actors \cite{chen2023interactive, chandra2022gameplan}.

% There has also been research in applying game-theoretic approaches for multi-agent planning with human drivers \cite{chandra2022gameplan}.

\begin{figure}
  \centering
  
  % --- Top row of images ---
  % We remove the \caption and \label from each subfigure
  % and add a simple text label below each image.
  % I've slightly increased the width for better spacing.
  \begin{subfigure}{0.15\textwidth}
    \includegraphics[width=\linewidth]{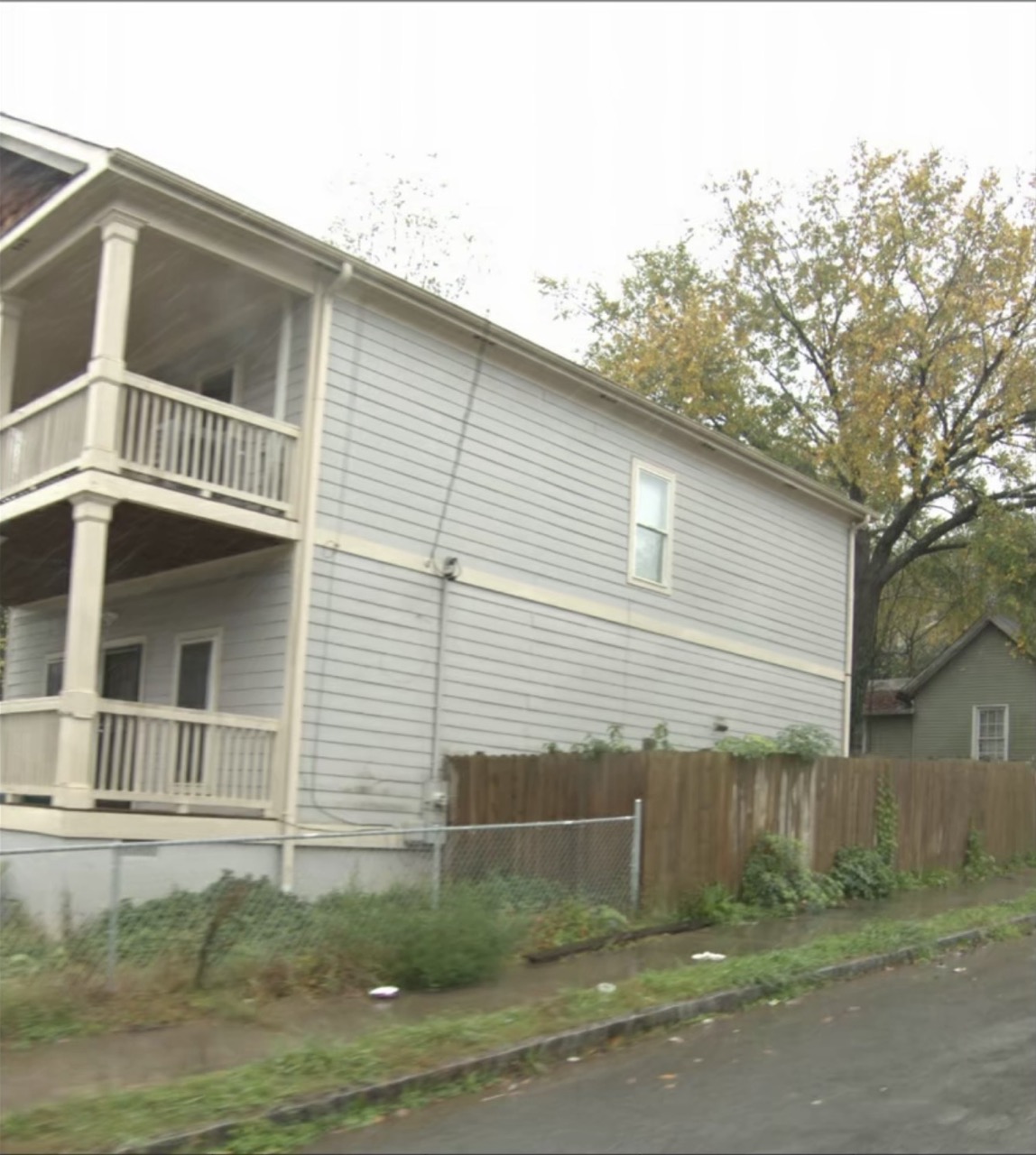}
    \centering (a)
  \end{subfigure}
  \hfill % This command distributes the horizontal space evenly
  \begin{subfigure}{0.15\textwidth}
    \includegraphics[width=\linewidth]{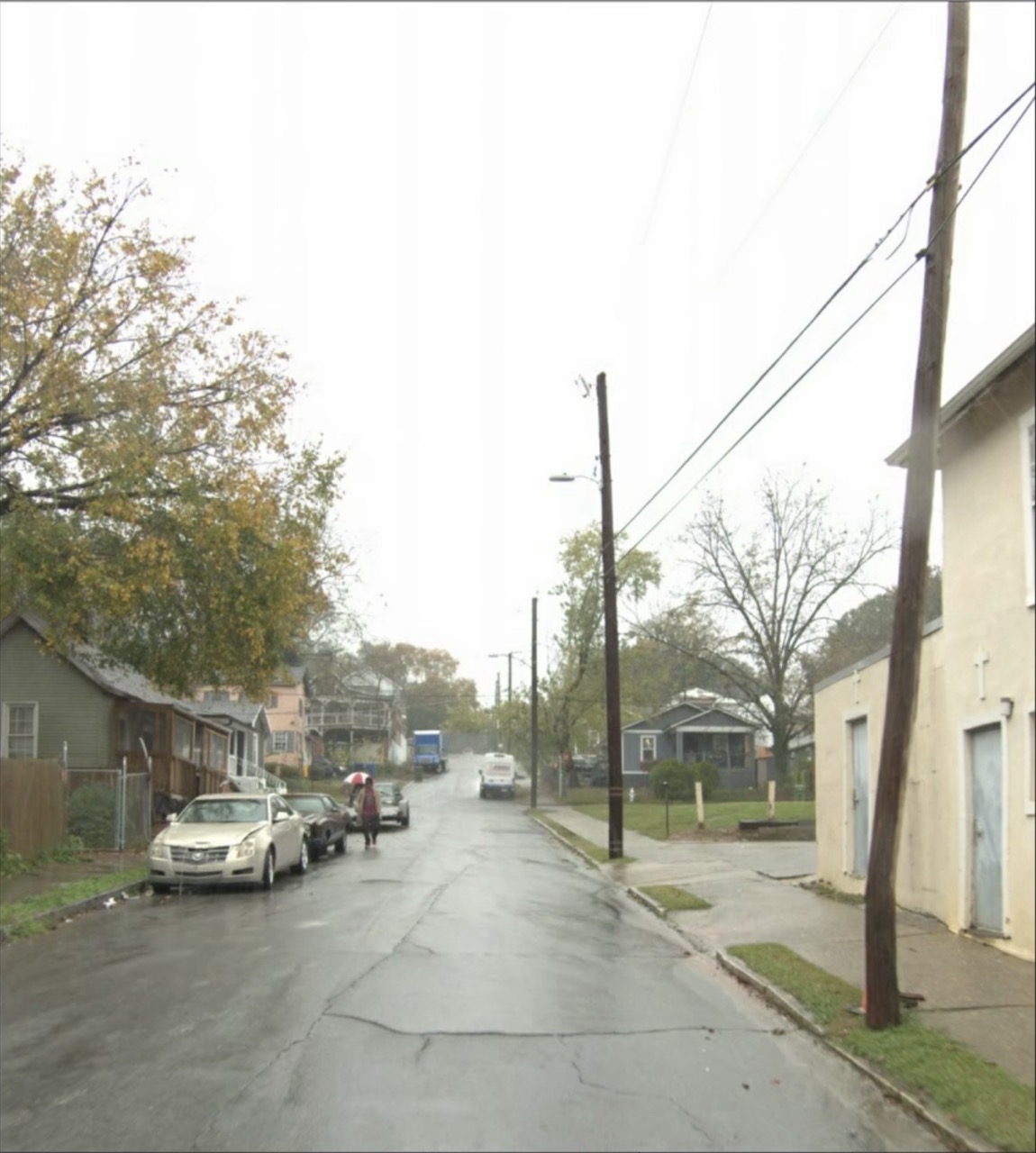}
    \centering (b)
  \end{subfigure}
  \hfill
  \begin{subfigure}{0.15\textwidth}
    \includegraphics[width=\linewidth]{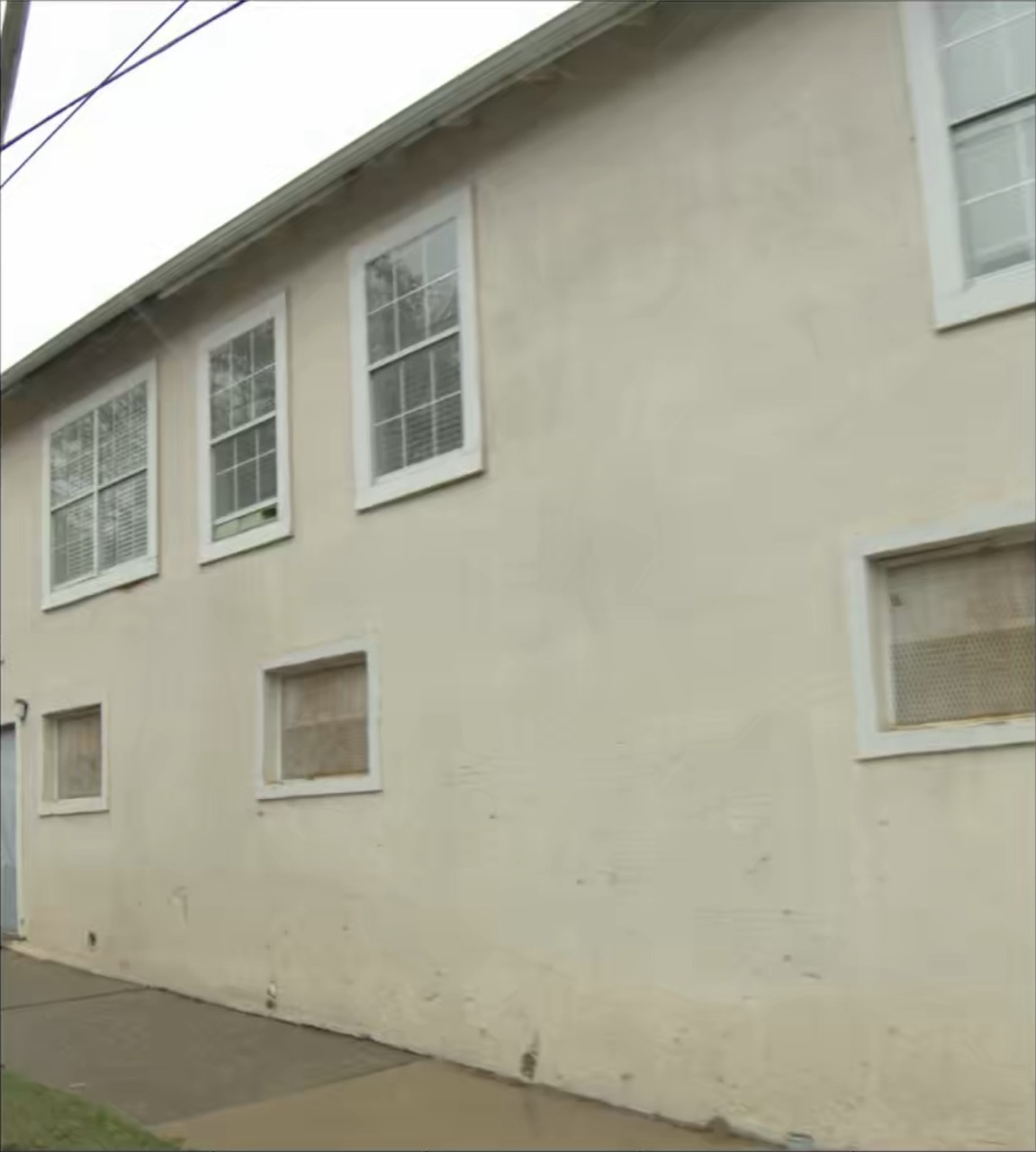}
    \centering (c)
  \end{subfigure}

  \vspace{1em} % Adds a little vertical space between the rows of images

  % --- Bottom row with the annotated image ---
  % I've made this subfigure wider to align better with the top row
  \begin{subfigure}{0.46\textwidth}
    \includegraphics[width=\linewidth]{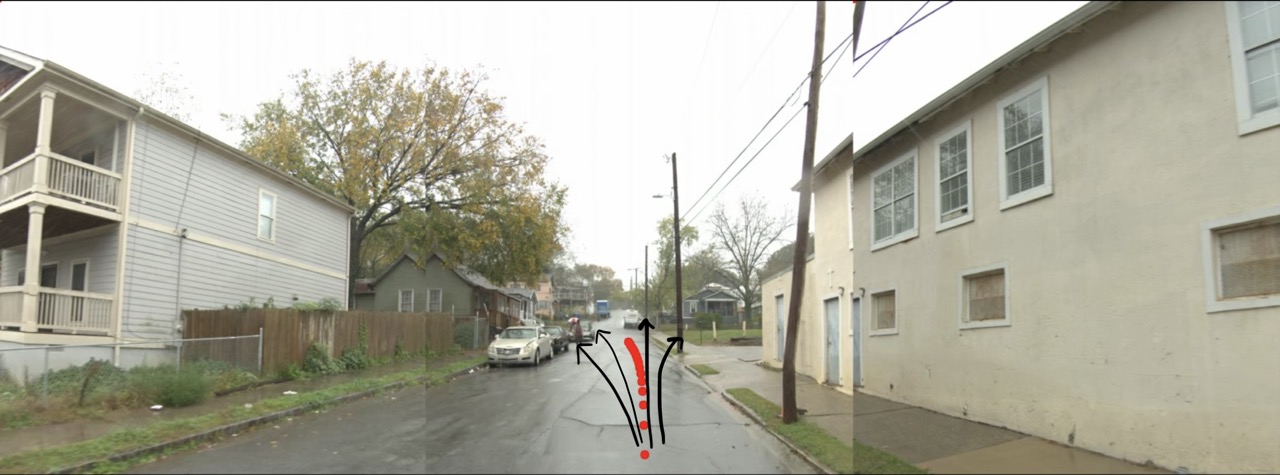}
    \centering (d)
  \end{subfigure}

  % --- The single, comprehensive main caption ---
  % We now define what (a), (b), (c), and (d) are right here.
  \caption{3 of the 8 camera images and trajectory annotations for a random sample from the Waymo dataset. (a) Front Left View, (b) Front View, (c) Front Right View. (d) The annotated front view showing the true trajectory in red and generated candidate trajectories in black. The intent for this sample is "1", which corresponds to going straight. The combination of scene input, vehicle history, and intent is used to generate the planned trajectory.}
  \label{fig:input_output}
\end{figure}

For the purposes of this paper, we distinguish between the autonomous driving prediction and planning tasks as follows: \textit{prediction} involves estimation of where an agent will move, or in the case of learning to model historical data, where an agent \textit{did} move. \textit{Planning}, on the other hand, is the generation of a feasible action sequence, conditioned on an intended goal. Different such plans may be equivalently optimal under some metric, or different plans may be individually optimal for different metrics; whereas prediction has only one correct answer, planning leaves open a variety of generative possibilities, which makes the evaluation of planning algorithms challenging. We provide an illustration of the input and expected output of our planning task in Figure \ref{fig:input_output}. 

Discussed in the following section, recent research has increasingly explored end-to-end (E2E) learning approaches toward both tasks \cite{hwang2024emma}, where raw sensor inputs are mapped directly to driving decisions. These methods are promising for handling complex environments and leveraging powerful representation learning from vision-language or vision-motion models, but broadly suffer from two challenges that reflect the long-tail problem:
\begin{enumerate}
    \item Models which output high-level control decisions, such as ``turn left" or ``go straight", oversimplify the driving problem to a decision which itself can have many instantiations which satisfy the meaning of the decision but do not conform to the scene, and
    \item Models which output a specific trajectory (or set of possible trajectories) are trained on prior data, which by definition of the long-tail tends to contain normal driving events and encounters, reducing model effectiveness when introduced to a novel scenario.
\end{enumerate}

A new dataset released by Waymo, named the End-to-End Driving Dataset\footnote{https://waymo.com/open/data/e2e}, provides a possible but underexplored solution to the second problem. In Waymo's words, this novel dataset contains 4,000 driving segments ``specifically curated to represent long-tail scenarios that drivers encounter in different environments, such as navigating construction zones during marathons, avoiding pedestrians falling off scooters, and maneuvering around unexpected obstacles on freeways. A mining analysis indicates that these events occur with a frequency of less than 0.003\% in daily driving, highlighting the dataset's unique focus on more rare and interesting scenarios.''

So, it is our hypothesis that, given appropriate long-tail data, an end-to-end model may learn fundamental planning behaviors that transcend normal driving encounters, allowing the model to function in both normal and novel driving scenarios by learning the behaviors human drivers take when encountering the unexpected. However, this requires a model of appropriate structure and size to learn relevant features and generate appropriate plans. Accordingly, in this research, we propose MTR-VP (Motion Transformer for Vision-based Planning), an architecture that integrates visual representations with trajectory prediction mechanisms. Our method leverages vision transformers for perception, transformers for temporal motion context, and a cross-attention-based decoder to fuse intent with learned scene understanding, improving decision-making in edge-case scenarios.

\section{Related Research}

The method presented in this research is a novel combination of elements from prior autonomous driving prediction and planning methods which each served a purpose for their specific task. Repeating what was mentioned in the introduction, and echoed by \cite{dauner2023parting, dauner2024navsim, hu2023planning, chen2024end}, there are fundamental misalignments between the goals of autonomous driving and the historical subtasks and respective evaluation metrics; generating a safe autonomous driving action requires more than learning to imitate past trajectories conditioned on agent positions. Semantic understanding of objects in the scene and their relationships and possible motions are important, whether learned implicitly by an end-to-end network or modularly for rule-based planners. Some methods even seek to explicitly model socially-compatible behaviors by reasoning about concepts like courtesy and prediction uncertainty \cite{wang2021socially}.

With respect to the above ideas, the method we present in this paper is end-to-end-\textit{like}. It is a method which implicitly learns to understand scene semantics for planning through attention mechanisms applied to the egocentric views, but at the same time, it does not output a single trajectory: it generates many, then classifies over these generated trajectories to select the plan with the highest likelihood given the scene and intention. Predicting multiple futures reduces the variance within each output mode and increases the probability of predicting the true trajectory, drawing from the multiple-trajectory-prediction principles introduced in \cite{cui2019multimodal, chai2020multipath, phan2020covernet, greer2021trajectory}, while adopting the neural-network-based egocentric scene understanding of end-to-end methods such as \cite{prakash2021multi, jia2023think, chitta2021neat, xu2024drivegpt4, xing2025openemma}. 

% This multi-modal approach aligns with research that aims to improve trajectory prediction by enforcing common-sense driving behaviors, such as conforming to the direction of a lane, through the use of auxiliary loss functions \cite{greer2021trajectory}

We especially draw from the representation learning of the Motion Transformer (MTR) method of \cite{shi2022motion, shi2024mtr++}, which has shown the efficacy of combining high-capacity deep models with rich spatial-temporal data. MTR combines map and actor features with multimodal intent-conditioned trajectory prediction. However, it depends heavily on high-definition (HD) maps and precomputed actor trajectories, which may not be available nor reliable in long-tail cases. Our model is inspired by the MTR framework, which included a scene context encoder and a decoder model. On a high level, their scene context encoder encodes the scene using map features, track information, and polylines available in the Waymo Motion dataset \cite{ettinger2021large}. In the MTP decoder, learnable intention queries help guide the ego vehicle in the direction it needs to move in for better alignment of future trajectory predictions with the goal direction. MTP employs cross attention between the intention queries and the scene context embeddings to produce the future trajectories.

Elaborated in the following sections, our approach draws inspiration from MTR but adapts it to a vision-first context, replacing map features with embeddings from a pretrained Vision Transformer (ViT) and modeling past kinematic states using a temporal transformer. There has been research that fuses the map information with the features from camera images using transformer-based methods \cite{jeong2024multi}, but here, we want to replace the map features with the camera features and the past kinetic data. Further, we apply the method to the end-to-end driving task, which is at its core a planning task, different than the prediction task which MTR is framed around, since the planning task includes an explicit intention or high-level goal for the ego vehicle.

\section{Problem Definition}

Here we describe the specific task addressed by our model. Given a 12-second history of vehicle state (position, velocity, acceleration), 8 surround-view camera images, and a routing intent, the task is to predict the future 5-second trajectory of the ego vehicle in bird's-eye view (BEV) coordinates at a 4Hz sampling rate (i.e. 20 (x, y) pairs representing a predicted trajectory). We note that this sits somewhere between ``prediction'' and ``planning'' as introduced in the earlier sections, for the following reasons:
\begin{itemize}
    \item The inclusion of an intention implies an agency that the generation must be conditioned on. The task is not simply guessing where a vehicle may go; it is knowing abstractly what the vehicle intends to do, then generating a series of waypoints which is conditioned on that intention and the observed scene.
    \item The method of evaluation leaves room for multiple ``correct'' plans, utilizing Waymo's Rater Feedback Score (RFS) as a primary planning metric, and a secondary metric of average displacement error (ADE) from the actually-taken trajectory as a prediction-centric proxy. Waymo's RFS compares predictions to three human-rated trajectories; generated trajectories falling within a dynamic trust region receive a higher rater's score, and others are penalized exponentially.
\end{itemize}

\section{Model Architecture}

\begin{figure*}[!t]
    \centering
    \includegraphics[width=\textwidth]{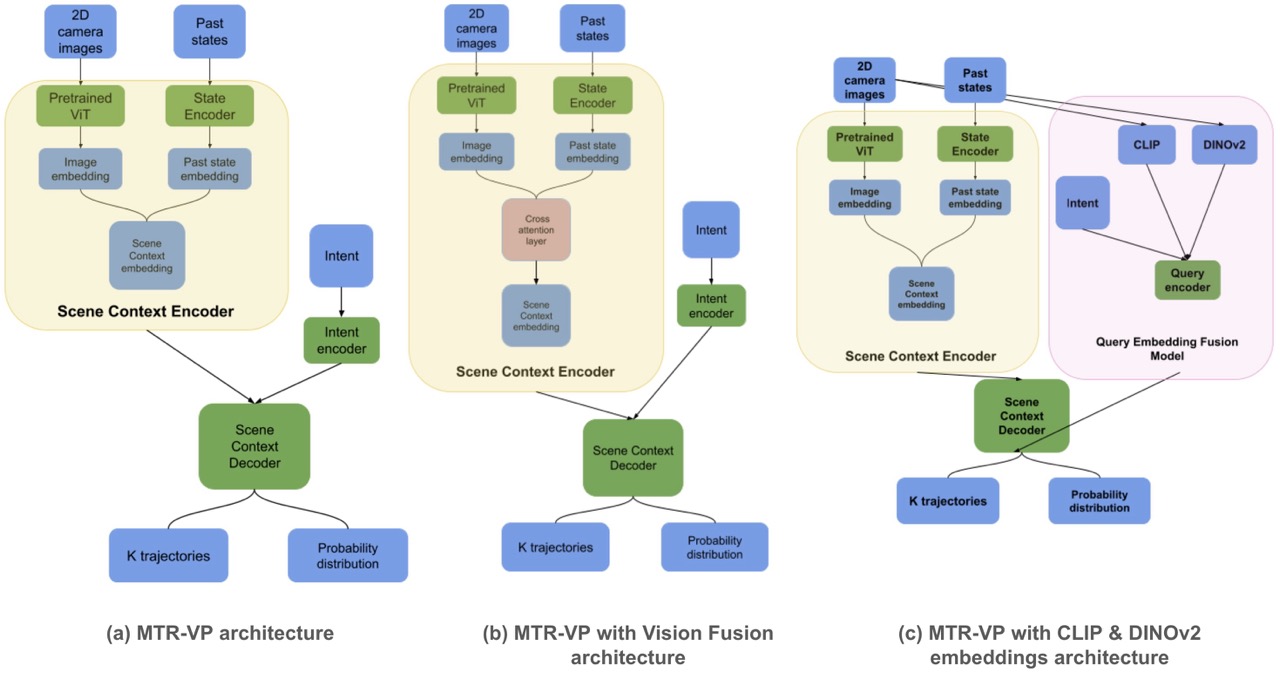}
    \caption{MTR-VP Model Architecture: the inputs, outputs, tensors, etc. are in blue blocks and the models are in green blocks. Camera images and past states are encoded and concatenated, then decoded to form multiple possible trajectories and an associated probability distribution (a). Cross attention is applied between an intent encoding and the scene context encoding in (b), shown in the pink box. In (c), cross attention is applied between the query embedding (generated with the intent, CLIP, and DINOv2 embeddings) and the scene context encoding, then decoded to output a single future trajectory.}
    \label{fig:model_architectures}
\end{figure*}

The input to our model includes the intent of the ego vehicle (i.e. some descriptor of where the ego vehicle aims to go in the future), 2D camera images, and past states of the ego vehicle (comprising of the positions, velocities, and accelerations). In our implementation, the discrete driving command associated with each frame simply indicates whether the intended route is towards the left, straight or right direction. True future states are made available for training, though we note that the RFS allows for additional trajectories which differ from the ground truth to be scored equally high. 

We take inspiration from the MTR framework to build a scene encoder transformer and a decoder to predict the future states. We follow this approach of a two-stage model (the scene context encoder and  decoder), but our model fuses the features obtained from the camera images with the past trajectory data unlike using the map features that the MTR model encoder uses. Unlike MTR, our approach does not use learnable intention queries since the Waymo End-to-End Driving dataset comprises of the intent. Our model architecture is shown in Figure 2 (a). The scene context encoder and the scene context decoder are shown in this part clearly.
%  from the Waymo Motion Dataset.

\subsection{Scene Context Encoder}

Unlike map-based approaches, our approach embeds the scene context from 2D camera images and past vehicle states.  In our particular implementation, we use three images: perspectives taken from the front view, the front left, and the front right. We stitch these images together to obtain one image that now looks like a panorama. This panorama is then rescaled to a size of 384 x 384 and passed into a pretrained Vision Transformer (ViT) \cite{dosovitskiy2020image}. ViT is pretrained on the ImageNet-21K dataset, a large-scale image classification dataset containing over 14 million images and 21,000 classes, which serves as a richer pretraining source than the commonly used ImageNet-1K. \cite{ridnik2021imagenet}. This dataset is a primary dataset used for training models for computer vision. ViT treats images as sequences of patches and applies attention on these patches to get rich embeddings of the entire scene. We use these embeddings as one part of the scene context embeddings.

The past states of the ego vehicle comprise of the past $t$ seconds worth of (x, y) positions, velocities, and accelerations of the ego vehicles, sampled at some rate $s$. In our case, we use $t=4$ seconds at $s=4$ Hz. We therefore have a sequence of 16 vectors: $\bar{v} = [x, y, v_x, v_y, z_x, z_y]^T$ where $x$ is the x-coordinate of the position, $y$ is the y-coordinate of the position, $v_x$ is the x-component of the velocity, $v_y$ is the y-component of the velocity, $z_x$ is the x-component of the acceleration, and $z_y$ is the y-component of the acceleration. We first encode this 6D vector into a state embedding and then use a transformer encoder model with 8 attention heads and 4 layers. Self attention in this block of the model helps encode the relationships between the past states, producing a rich embedding of the vehicle's recent history. 

The image embeddings and the state embeddings are concatenated to form the final scene context embeddings.

\subsection{Scene Context Decoder}

The intent and the scene context embeddings obtained from the scene context encoder are used as inputs to the scene context decoder. 

For use in the decoder, the intent is first encoded using a linear layer. In the dataset used in this paper's experiments, the intent is provided as an integer value; 1 for going straight, 2 for going left, and 3 for going right. The intent embedding layer is a linear layer that encodes the intent value.

We employ cross attention between the intent embeddings and the scene context embeddings to learn the relation between the intent values and the past states of the ego vehicles. The intent embedding is used as the query while the scene context embeddings are used as the keys and the values. Intuitively, this translates to assigning the correct importance to the past state and scene embedding based on the intent of the ego vehicle as it navigates its way on the street. These rich embeddings obtained from the cross attention layers are fed to a fully connected network that translates these to two sets of output: K possible future trajectories along with the probability distribution for these trajectories. 

% Intuitively, this translates to paying attention to which features of the past state embedding and the scene embedding depend on the intent of the ego vehicle as it navigates its way on the street. 

\subsection{MTR-VP with Vision Fusion}
In this approach, we make a  modification to the previous model. The concatenation of the image embeddings on top of the past trajectory embeddings does not integrate the visual embeddings effectively along with the past trajectory embeddings in the scene context embeddings. This is clear from the results of our ablation studies where even when the model is provided with blank images, the ADE metric values are very similar. So, we make a change to the scene context encoder and instead of concatenating the visual embeddings on top of the past trajectory embeddings, we use another multi-head cross attention module where we use the sequence of the past trajectory embeddings as the queries and the visual embeddings as the keys and values. This architecture for this approach is shown in Figure 2(b).

\subsection{MTR-VP with Foundation Model Embeddings}

The decoder in the MTR-VP model is very simplistic where we try to use the intent embedding as the query since we want to generate a future trajectory for that specific route intent. However, we must consider that along with the direction that we want to take, the model must also be aware of various factors about the road in the camera images like the curvature of the road, the construction zones, the lanes, etc. So, this must be part of the query as well along with the route intent. We added CLIP and DINOv2 embeddings of the camera images to the image embeddings while creating the queries in the cross attention step of the transformer, to provide the model the opportunity to exploit representations learned in these large-scale foundation models which may be more rapidly distilled to driving decisions than the information from the accompanying full-scale images. The CLIP embeddings can provide language-related scene semantics which may be useful context in navigating the scene while also following the encoded intent in cases like lane curvatures, lane closures, traffic information, etc. When these features are included in the query, the model is predicting the future trajectory conditioned on the route intent and also the linguistic description of the scene. The DINOv2 embeddings provide very rich embeddings for visual context, and we include these embeddings to enhance the query embeddings in the decoder. Figure 2(c) shows the architecture of this approach. We used a fusion model which is a 2-layer FCN on the concatenation of the intent embedding, the CLIP embedding, and the DINOv2 embeddings to get the final query embeddings. This embedding contains the information about the scene along with the intent which can help the model predict the future trajectory according to the route intent and the scene information.

\subsection{Training}

For each training data sample, we know the ego vehicle's future 5-second trajectory, in the same format and frequency as the past trajectory. From the ViT model, we use the 768-dimensional embedding for the CLS token (also called the ``pooler'' output), which is a rich embedding of the entire image. For the state encoders, we use 8 attention heads and 4 layers. The query, key, and value weight matrices thus have the shape 6 x 64. The output of the state encoder is $B \times 16 \times 768$. Here $B$ is the batch size and 16 is the total length of the sequence of the past states. We combine these to get the scene context embeddings of shape $B \times 17 \times 768$. 

In the scene context decoder, we encode the intent into a 128 dimensional vector. In the cross attention, the query weight has the shape $128 \times 512$, while the key and value weight matrices have the shape $768 \times 512$ each. The output of the cross attention is fed to a fully connectd layer with a single linear layer that produces an output of shape $B \times (K \times 20 \times 2)$, where $K$ is the number of predicted trajectories, 20 is the length of the sequence of future states. We predict $K=20$ trajectories. This is reshaped to get an output of shape $B \times K \times 20 \times 2$.

We use a combination of the cross entropy loss and the L2 regression loss to train our model. We use the cross entropy loss to increase the probability for the trajectory that is closest to the actually-driven future trajectory. We are deliberately shying away from the term ``ground truth'' in this instance, as the path driven in reality may not be the only acceptable planned path towards a particular goal. The L2 regression loss is used to bring only the closest future trajectory prediction closer to the target trajectory. 

We use the Adam optimizer for training. It adapts learning rates for each parameter using estimates of first and second moments of gradients, enabling faster convergence and better performance on sparse or noisy data. We train our model on 8 NVIDIA L40 24 GB GPUs. 

Waymo provides a training data partition of 2,037 20-second driving segments, which we further divide into 80\% training data and 20\% validation data. As will be discussed in later sections, we utilize Waymo's separately-designated validation data partition of 479 20-second driving segments as a ``test'' set, because Waymo's designated test set of 1,505 12-second driving segments does not come with any labels, as it is meant for private evaluation on a server.

\section{Experimental Evaluation}

\subsection{Dataset}

The Waymo End-to-End Driving Dataset employs two primary coordinate systems: vehicle coordinates and sensor frames. The vehicle coordinate system is centered at the ego vehicle's center, with the x-axis pointing forward, the y-axis pointing left, and the z-axis pointing upward. All trajectory data is referenced to this vehicle coordinate system. Each sensor's frame is related to the vehicle frame by an extrinsic transformation. For cameras, the right-handed coordinate system frame is centered at the lens. The x-axis points out from the lens, the z-axis points upward, and the y/z plane is parallel to the camera's image plane. 

The dataset includes images from eight cameras, each capturing a different direction: front, front left, front right, side left, side right, rear, rear left, and rear right. We utilize three of these camera views on our method. For each direction, a single JPEG image is provided. Camera intrinsics and extrinsics are available to define each camera's position relative to the vehicle's center, enabling projection of 3D trajectories onto the camera images. Each driving segment features 10Hz camera video sequences, with training data spanning 20 seconds and testing data spanning 12 seconds.
We make the code for all our experiments and models publicly available at [anonymized] for future research. We note that, at the time of writing, our research is the only open-sourced model and experiments built around this dataset.

\subsection{Metrics}

The Rater Feedback Score (RFS) is defined and scored at 3 and 5 seconds into the future, and these scores are averaged over 11 different types of scenarios, including construction zones,
   intersections,
    pedestrian and cyclist interaction scenarios,
   single and multi-lane maneuvers, 
    cut-ins,
    foreign object debris scenarios, and
    special vehicle scenarios.

The RFS is a metric that compares a predicted trajectory against the closest rater-scored trajectory, where expert human raters provided a score of 0 (worst) to 10 (best) for 3 trajectory proposals for a given scene, intended to capture the diversity of acceptable driving decisions during critical events. In each sample, at least one of the rater specified trajectories has a score higher than 6.

For the purposes of computing RFS, a ``trust region" is defined as the area within specified lateral and longitudinal thresholds relative to the rater-specified trajectory at a given time point $t$, in this case specifically evaluated at $t=3$s and $t=5$s. We use the base thresholds for these trust regions adopted by Waymo: a lateral threshold $\eta_{\text{lat}}(3) = 1.0$ and longitudinal threshold $\eta_{\text{long}}(3) = 4.0$ at $t = 3$ seconds, and a lateral threshold $\eta_{\text{lat}}(5) = 1.8$ and longitudinal threshold $\eta_{\text{long}}(5) = 7.2$ at $t = 5$ seconds.

As described on the Waymo End-to-End Driving Challenge page\footnote{https://waymo.com/open/challenges/2025/e2e-driving/}, these thresholds are further scaled by the initial speed ($v$) of the rater specified trajectory using a piecewise linear function of the initial speed $v$ (in m/s):

\begin{equation}
    \text{scale}(v) = \begin{cases} 0.5, & v < 1.4 \text{ m/s} \\ 0.5 + 0.5 \times \frac{v - 1.4}{11 - 1.4}, & 1.4 \text{ m/s} \le v < 11 \text{ m/s} \\ 1, & v \ge 11 \text{ m/s} \end{cases}
\end{equation}

The final thresholds used to define the trust region at a given time $T$ (3s or 5s) and initial speed $v$ are then determined by multiplying the base thresholds by the calculated scale factor:

\begin{equation}
    \eta_{\text{lat}}(T, v) := \text{scale}(v) \times \eta_{\text{lat}}(T)
\end{equation}
\begin{equation}
    \eta_{\text{long}}(T, v) := \text{scale}(v) \times \eta_{\text{long}}(T) 
\end{equation}

For a predicted trajectory that falls within a trust region, the score assigned is the score of the corresponding closest rater specified trajectory (adjusted distance). If a predicted trajectory is outside of any trust regions, a score is assigned that is exponentially lower than the score of the closest rater specified trajectory, denoted as $\hat{x}$. This score depends on the distance error $\Delta$ between the predicted trajectory and the closest rater trajectory. The formula for assigning the score in this case is:

\begin{equation}
    \text{RFS} = \hat{x} \times 0.1^{\min(\max(\Delta - 0.5, 0), 1.0)}
\end{equation} 

Furthermore, for predicted trajectories that are outside trust regions, a floor score of 4 is assigned.

Additionally, we evaluate performance on the standard Average Displacement Error metric, which takes the L2 distance between each corresponding waypoint in the actual and predicted trajectory. This provides a measure of the similarity of the learned plan to a feasible prior plan. We measure this value at 3 seconds and 5 seconds after the time of interest.

\subsection{Results}

% Results guide on waymo website:
% MTR-VP with blank images : 8/7/2025 8:31 PM
% MTR-VP with single trajectory : 8/8/2025 0:28 AM
% MTR-VP with Vision Fusion with blank images : 8/8/2025 4:33 AM
% MTR-VP with Vision Fusion with single trajectory : 8/9/2025 1:05 AM
% MTR-VP with Vision Fusion : 8/7/2025 2:42 AM
% MTR-VP with CLIP & DINOv2 : 10/10/2025 7:58 PM
% MTR-VP with CLIP & DINOv2 with blank images : 10/10/2025 9:27 PM

\begin{table*}
\centering
\caption{Average Displacement Error (ADE) for different prediction horizons across methods on the test split.}
\begin{tabular}{|l|c|c|c|c|c|c|c|}
\hline
\multirow{2}{*}{Model} & \multicolumn{1}{c|}{3 second trajectory} & \multicolumn{1}{c|}{5 second trajectory} \\
 & ADE top-1 & ADE top-1  \\
\hline
UniPlan & \textbf{1.2671} & \textbf{2.8423} \\
DiffusionLTF & 1.3561 & 2.8914 \\
Waymo Baseline\footnote{No paper available.} & 3.0182  & 3.0182  \\
AutoVLA\footnote{Method from UCLA Mobility Lab; no paper available.} & 1.3507 &  2.9580 \\

OpenEMMA \cite{xing2025openemma} &  6.6842 & 12.4755  \\

\textbf{MTR-VP (ours)} &  1.4232 & 3.3485 \\

MTR-VP with Vision Fusion (ours) & 1.5451 & 3.5297 \\
MTR-VP with CLIP \& DINOv2 embeddings & 2.0142 & 4.3379 \\

MTR-VP with blank images (ablation study) & 1.4238 & 3.3509 \\
MTP-VP with single trajectory (ablation study) & 1.3504 & 3.0526 \\
MTR-VP with Vision Fusion with blank images (ablation study) & 1.5448 & 3.528 \\
MTR-VP with Vision Fusion with single trajectory (ablation study) & 1.3131 & 2.9835 \\

\bottomrule
\end{tabular}
\label{tab:ade_metrics}
\end{table*}

\begin{table*}
\centering
\caption{Average Displacement Error (ADE) of MTR-VP for different prediction horizons and top-K predictions on the validation split.}
\begin{tabular}{|c|c|c|c|c|c|c|}
\hline
{} & \multicolumn{3}{|c|}{3 second trajectory} & \multicolumn{3}{c|}{5 second trajectory} \\
\cline{1-7}
Model & ADE top-1 & ADE top-5 & ADE top-10 & ADE top-1 & ADE top-5 & ADE top-10 \\
\hline
MTR-VP & \textbf{0.6762} & \textbf{0.4117} & \textbf{0.3204} & \textbf{1.7680} & \textbf{0.9639} & \textbf{0.7087} \\
\hline
MTR-VP with Vision Fusion & 0.8071 & 0.4503 & 0.3446 & 1.9510 & 1.0458 & 0.7438 \\
\bottomrule
\end{tabular}
\label{tab:ade_metrics_different_k}
\end{table*}

\begin{table*}
\centering
\caption{Rater Feedback Score (RFS)  across methods on Waymo Test split for various categories.}
\begin{tabular}{|c|c|c|c|c|c|c|c|}
\hline
Model & Overall &  Construction & Intersection &  Pedestrian & Cyclist  \\
\hline
UniPlan & \textbf{7.7795} & \textbf{8.5600} & 7.8639 & 7.6384 & 7.7559 \\
DiffusionLTF & 7.7172 & 8.2601 & \textbf{7.9269} & \textbf{7.9085} & \textbf{7.7965} \\
Waymo Baseline &  7.5281  & 8.2729 & 7.6226 & 7.6651 & 7.5172  \\

AutoVLA &  7.5566 & 7.9556 &  7.7112 & 7.5920 & 7.3208 \\

OpenEMMA \cite{xing2025openemma}  &  5.1575 & 4.6574 & 5.4567 & 5.6003 & 4.9514  \\

\textbf{MTR-VP (ours)} &  7.3433 & 7.7111 &  7.6864 & 7.5096 & 7.3138 \\

\textbf{MTR-VP with Vision Fusion (ours)} & 7.3096 & 7.5359 & 7.5258 & 7.5895 & 7.0948 \\
\textbf{MTR-VP with CLIP \& DINOv2 embedidngs (ours)} & 6.7862 & 6.9871 & 6.7769 & 7.1561 & 7.0224 \\ 

MTR-VP with blank images (ablation study) & 7.3409 & 7.7116 & 7.6867 & 7.5098 & 7.3141 \\
MTP-VP with single trajectory (ablation study) & 7.2875 & 8.1392 & 7.4237 & 7.3986 & 7.3608 \\
MTR-VP with Vision Fusion with blank images (ablation study) & 7.3057 & 7.5342 & 7.5255 & 7.5892 & 7.0951 \\
MTR-VP with Vision Fusion with single trajectory (ablation study) & 7.3189 & 8.0359 & 7.5974 & 7.3863 & 7.2177 \\

\bottomrule
\end{tabular}
\label{tab:rfs_metrics1}
\end{table*}

\begin{table*}
\centering
\caption{Rater Feedback Score (RFS)  across methods on Waymo Test split for additional categories.}
\begin{tabular}{|c|c|c|c|c|c|}
\hline
Model & Cut-In &  Special Vehicles & Single Lane &  Multi-Lane & Debris \\
\hline
UniPlan & 7.7859 & 7.6702 & 8.1599 & \textbf{7.6699} & 8.0847 \\
DiffusionLTF & 7.6933 & 7.7401 & \textbf{8.2603} & 7.4163 & \textbf{8.0938} \\
Waymo Baseline &  \textbf{7.8690}  & 7.4597 & 7.8515 & 7.3012 & 7.7627 \\

AutoVLA &  7.5256 & 7.6968 & 8.1450 & 7.5100 & 7.9074 \\

OpenEMMA \cite{xing2025openemma}  &  5.1392 & 5.6639 & 5.3269 & 5.2768 & 4.6742  \\

\textbf{MTR-VP (ours)} &  7.4170 & 7.5705 &  7.7170 & 7.2205 & 7.3794 \\

\textbf{MTR-VP with Vision Fusion (ours)} & 7.5192 & \textbf{7.8479} & 7.5804 & 7.2827 & 7.4397 \\
\textbf{MTR-VP with CLIP \& DINOv2 embedidngs (ours)} & 6.9529 & 6.8868 & 7.5726 & 6.3238 & 6.9512 \\ 

MTR-VP with blank images (ablation study) & 7.4050 & 7.5845 & 7.7169 & 7.2211 & 7.3798 \\
MTP-VP with single trajectory (ablation study) & 7.5030 & 7.3308 & 7.5358 & 7.0508 & 7.4937 \\
MTR-VP with Vision Fusion with blank images (ablation study) & 7.4787 & 7.8477 & 7.5798 & 7.2830 & 7.4399 \\
MTR-VP with Vision Fusion with single trajectory (ablation study) & 7.4345 & 7.4024 & 7.6766 & 6.8489 & 7.6465 \\

\bottomrule
\end{tabular}
\label{tab:rfs_metrics2}
\end{table*}

\begin{figure*}
  \centering
  \begin{subfigure}{0.32\textwidth}
    \includegraphics[width=\linewidth]{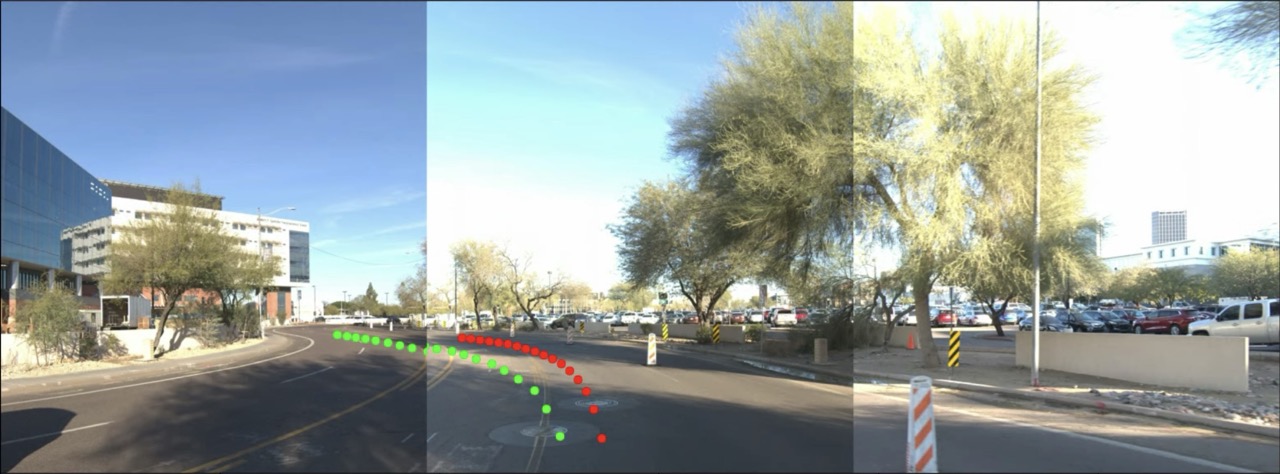}
    \caption{Traffic Barriers}
    \label{fig:front_left}
  \end{subfigure}
  % \hfill
  \begin{subfigure}{0.32\textwidth}
    \includegraphics[width=\linewidth]{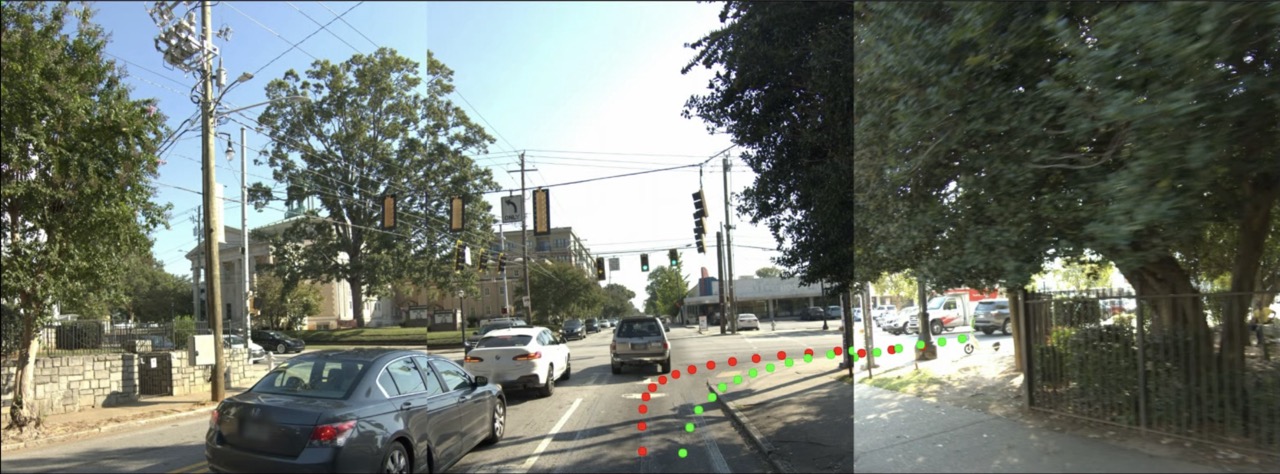}
    \caption{Unmarked Right}
    \label{fig:front}
  \end{subfigure}
  % \hfill
  \begin{subfigure}{0.32\textwidth}
    \includegraphics[width=\linewidth]{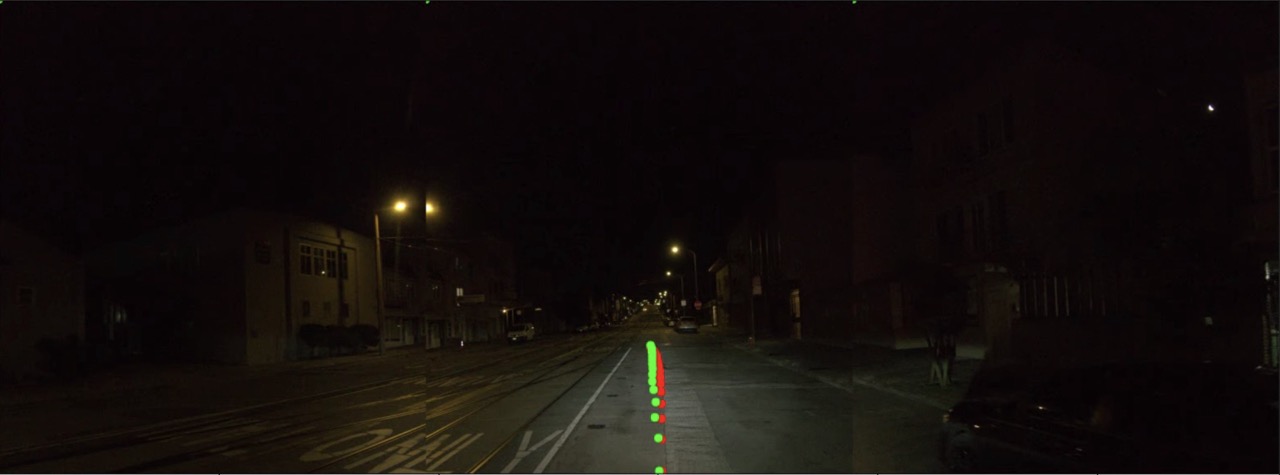}
    \caption{Night-time}
    \label{fig:front_right}
  \end{subfigure}
  \begin{subfigure}{0.32\textwidth}
    \includegraphics[width=\linewidth]{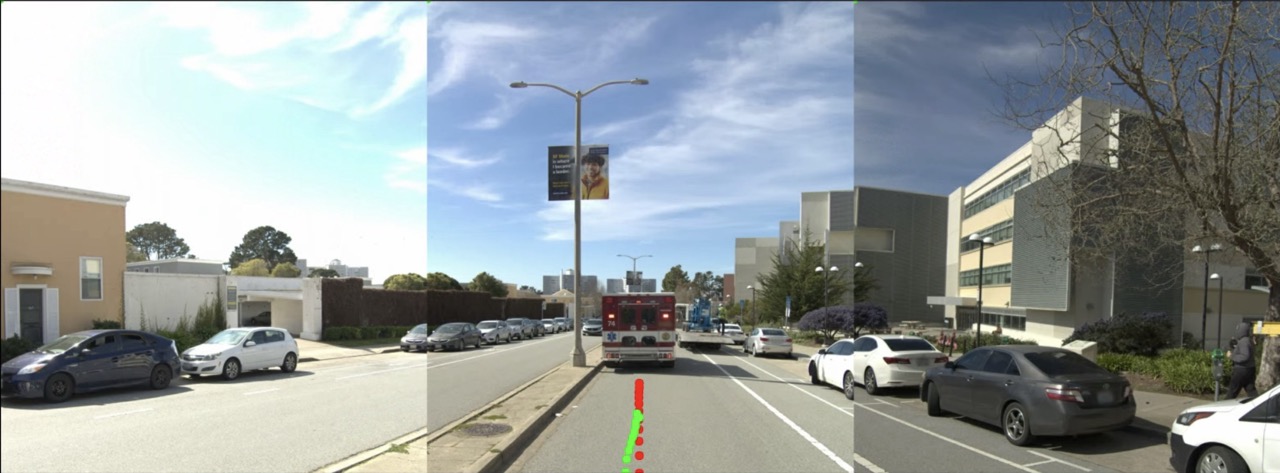}
    \caption{City Bus}
    \label{fig:front_right}
  \end{subfigure}
  \begin{subfigure}{0.32\textwidth}
    \includegraphics[width=\linewidth]{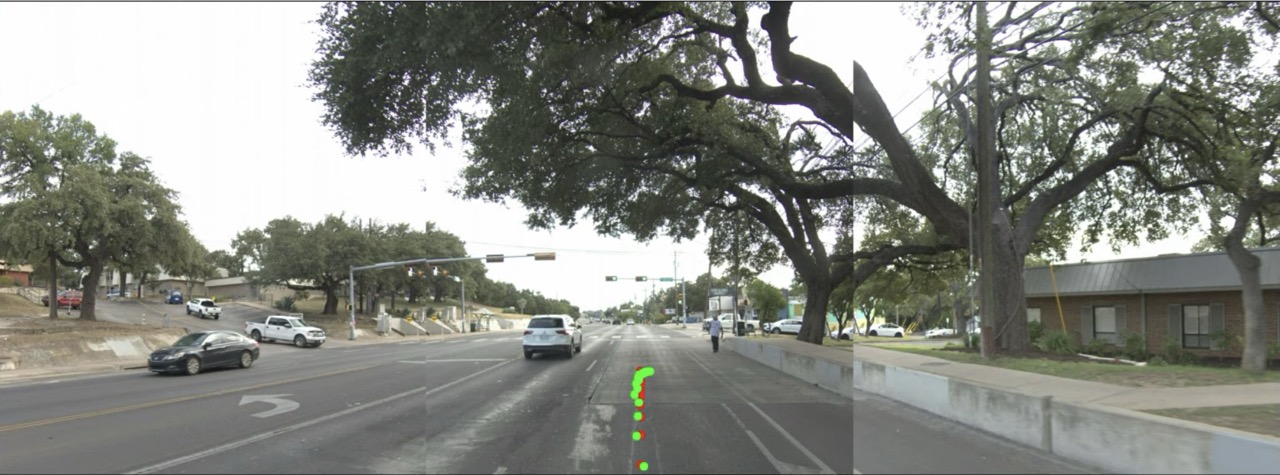}
    \caption{Pedestrian}
    \label{fig:front_right}
  \end{subfigure}
  \begin{subfigure}{0.32\textwidth}
    \includegraphics[width=\linewidth]{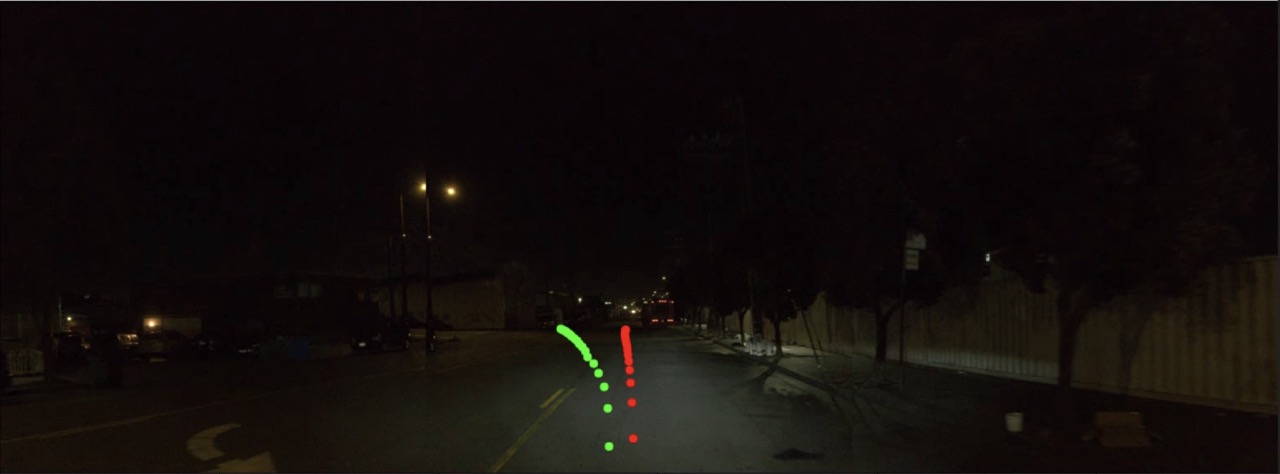}
    \caption{Night-time Left}
    \label{fig:front_right}
  \end{subfigure}

  \caption{Predicted vs. ground-truth trajectories overlaid on each of the front-facing camera views. Red dots indicate actual trajectories, and green dots indicate model predictions. The highest-likelihood model predictions are very similar to actual trajectories taken.}
  \label{fig:camera_views}
\end{figure*}

We evaluate our model on the Waymo End-to-End Driving Dataset validation split, because the dataset test split is without annotations for evaluation. However, test split performance is available via a reporting server, which we utilize to create Table \ref{tab:ade_metrics}. We note that even though the Waymo Validation split and Waymo Test split are different data splits, neither are used in any training or hyperparameter selection for MTR-VP, providing the next-best means of comparison to existing methods. Table \ref{tab:ade_metrics_different_k} illustrates the effectiveness of the multiple-trajectory generation; as the number of generated trajectories considered in evaluation increases, the ADE decreases, showing that good trajectories are being generated even though the model may occasionally struggle to select the best fit. In Figure \ref{fig:camera_views}, we visualize predictions in select complex scenes to qualitatively illustrate  the model's response to long-tail events.

The models were trained to minimize the ADE metric. We expect the models trained on this objective function to also do well on the RFS metric. This is because the model is expected to pick up on visual as well as past trajectory features useful for predicting the future trajectory. 

We have presented our results on the ADE metric for a 3-second trajectory as well as a 5-second trajectory for all the models we have trained in Table I, including ablations. The RFS metrics for all the various scenario categories for our trained models and ablations are presented in Tables III and IV. 

From Table I, we can see that the ADE metrics for the 3-second trajectory as well as the 5-second trajectory are very similar for the MTR-VP model and the MTR-VP model evaluated with blank images. We see that this is the case for the MTR-VP with Vision Fusion model as well. This means that the visual embeddings are not being combined properly with the past trajectory information using the transformer methods to produce useful and rich scene context embeddings; in other words,  our models are lacking `vision` and learning a purely past trajectory-driven prediction task. This would mean that the model predicts incorrect future trajectories when there are lane curvatures, construction zones, lane closures, etc. This is important to understand because taking these factors into consideration is of paramount importance for the safety factor in autonomous driving. 

However, from Tables I and III, we can see that in spite of the image information not being factored in, our ADE metrics are actually better than the Waymo baseline model, and the RFS metrics are comparable. This means that the transformer-based methods have the potential to find rich features from the past trajectory data to predict future trajectories. If the visual features are effectively combined with the past trajectory features obtained from the transformer-based encoder model, then there will be a high scope for improvement in future trajectory prediction models.

In the second ablation study, we assessed whether the multi-trajectory-prediction approach helps with the future trajectory prediction. From Tables I and III, we can see that our multi-trajectory-prediction approach for the MTR-VP and MTR-VP with Vision Fusion models resulted in better ADE and RFS metrics than the single-trajectory-prediction approach applied to the respective models. This makes sense because when we predict multiple possible trajectories along with the probabilities for each of those modes, we reduce the variance within each `mode' and the model learns to predict the most probable future trajectory by first evaluating multiple possible futures. 

We also experimented with enhanced query embeddings for nuanced cases where the vehicle must condition on factors like lane closures, traffic, etc. along with the route intent while predicting the future trajectory, but from our results in Tables I, III, and IV, we see that the MTR-VP with CLIP \& DINOv2 embeddings model performed  worse than our MTR-VP and MTR-VP with Vision Fusion models. Language-driven embeddings describing the scene would ideally be beneficial to the model so that there is one more layer of visibility that the model can use for predicting safe future trajectories, however, our method of combining these features seem to be insufficient, possibly due to the creation of a higher-dimensional input space relative to the volume of training data available, or shortcomings of the ability of these foundation models to deal with spatial grounding of features within the visual scene. Utilizing these features in the query embeddings can be prove to be great future direction for research. 

\section{Conclusion}

In this research, we introduce MTR-VP, a vision-first adaptation of the Motion Transformer framework for trajectory planning in rare and challenging driving scenarios. We see that transformer-based methods are good at extracting useful and rich features from sequential kinetic data from past trajectory information, however, they are not as effective at combining these features with the visual features that embed the information about the scene, supported by our ablations of visual input. Our results highlight the scope for exploring ways of combining the rich past trajectory features with the visual features encoded in the camera images to produce better and safer future trajectory predictions, especially in the application of foundation model embeddings of visual input. 

Novelly, we show that multi-trajectory-output, a technique effective in imitation learning of agent futures, is also beneficial in the future trajectory planning task, allowing the model to explore and select from multiple trajectory modes for a given scene. 

While this research represents a novel and high-performing method towards trajectory planning in autonomous driving, there exist limitations to both the method and the framing of the problem which should be overcome to create more robust autonomous driving intelligence. First, there is a limited ability of current metrics to effectively evaluate the desired algorithmic behavior. While alignment to expert human ratings is one way to measure performance, this leaves many gaps in subjectivity and safety, and further, limiting to three candidate trajectories fails to capture the true breadth of feasible decisions available to the vehicle. Second, this method is applicable only in short-horizon single-instruction maneuvers, such as turning left, turning right, or going straight. The ability of this method to generate trajectories for more complex, composite maneuvers is a natural next step, so that autonomous vehicles can make higher-level plans which may depend on a longer temporal context. In future work, integrating LLM-based intent priors may lead to richer routing understanding, which may further address long-tail scenarios faced in real-world driving.

\bibliographystyle{IEEEtran}
\bibliography{refs}

% %%%%%%%%%%%%%%%%%%%%%%%%%%%%%%%%%%%%%%%%%%%%%%%%%%%%%%%%%%%%

% \appendix
% %%%%%%%%%%%%%%%%%%%%%%%%%%%%%%%%%%%%%%%%%%%%%%%%%%%%%%%%%%%%

\end{document}